\documentclass[UTF8]{article}

    \PassOptionsToPackage{numbers, compress}{natbib}

\usepackage[preprint]{Query-Kontext}

\usepackage[utf8]{inputenc} 
\usepackage[T1]{fontenc}    
\usepackage{hyperref}       
\usepackage{url}            
\usepackage{wrapfig,lipsum}
\usepackage{booktabs}       
\usepackage{amsfonts}       
\usepackage{nicefrac}       
\usepackage{microtype}      
\usepackage{xcolor}         
\usepackage{booktabs}

\usepackage{caption}
\usepackage{subcaption}
\usepackage{bbding}
\usepackage{graphicx}
\usepackage{multirow}

\AtEndPreamble{
    \usepackage[capitalize]{cleveref}
    \crefname{section}{Sec.}{Secs.}
    \Crefname{section}{Section}{Sections}
    \Crefname{table}{Table}{Tables}
    \crefname{table}{Tab.}{Tabs.}
}

\usepackage{colortbl}
\definecolor{mygray}{gray}{0.95}
\definecolor{my_green}{RGB}{82,208,80}
\usepackage{enumerate}
\usepackage{makecell}
\definecolor{00red}{RGB}{236,35,35}

\usepackage{pifont}

\definecolor{mygreen}{RGB}{112, 180, 143}
\definecolor{myred}{RGB}{242, 128, 128}

\definecolor{lightgray}{gray}{.9}
\definecolor{lightblue}{RGB}{230,240,255}
\definecolor{lightgreen}{RGB}{230,255,230}
\definecolor{lightyellow}{RGB}{255,255,230}
\definecolor{lightred}{RGB}{255,230,230}

\definecolor{lightlightgray}{gray}{.95}
\definecolor{lightlightblue}{RGB}{240,245,255}
\definecolor{lightlightgreen}{RGB}{240,255,240}
\definecolor{lightlightyellow}{RGB}{255,255,240}
\definecolor{lightlightred}{RGB}{255,240,240}

\definecolor{lightlightlightgray}{gray}{.99}
\definecolor{lightlightlightblue}{RGB}{247,250,255}
\definecolor{lightlightlightgreen}{RGB}{247,255,247}
\definecolor{lightlightlightyellow}{RGB}{255,255,247}
\definecolor{lightlightlightred}{RGB}{255,247,247}

\newcommand{\cmark}{\textcolor{mygreen}{\ding{51}}}%
\newcommand{\xmark}{\textcolor{myred}{\ding{55}}}%

\newcommand{\rarrow}{\raisebox{0.6\height}{\,$\rightarrow$\,} }

\usepackage[normalem]{ulem}
\usepackage{comment}

\newcommand{\ours}{Query-Kontext}

\usepackage{amsmath}
\usepackage{fancyhdr}

\usepackage{tabularx}
\usepackage{adjustbox}

\usepackage[most]{tcolorbox}
\definecolor{BoxBackground}{RGB}{240, 240, 240} 
\definecolor{BoxFrame}{RGB}{0, 0, 0} 
\definecolor{TitleBackground}{RGB}{0, 0, 0} 
\definecolor{TitleText}{RGB}{255, 255, 255} 

\tcbset{
  academicbox/.style={
    boxsep=5pt,
    left=2pt,
    right=2pt,
    bottom=0.5pt,
    boxrule=0.5pt,
    colback=BoxBackground,
    colframe=BoxFrame,
    colbacktitle=TitleBackground,
    coltitle=TitleText,
    enhanced,
    attach boxed title to top left={yshift=-0.1in,xshift=0.1in},
    boxed title style={boxrule=0pt,colframe=white},
    title={#1},
  }
}
\newtcolorbox{AcademicBox}[1][]{academicbox=#1}

\title{Decoupling the Multi-Modal Reasoning via Query-Kontext and Scaling the Image Diffusion for Unified Image Generation and Editing.}
\title{Query-Kontext / Kontext-Guidance: Unlashing Multi-modal generative reasoning from Vision-language Model for Unified Image Generation and Editing}
\title{Unlashing, Scaling and Decoupling for Unified Multi-modal to Image Model}
\title{Unlashing, Scaling and Decoupling Unified Multimodal Models for Image Generation and Editing}
\title{Query-Kontext: An Unified Multimodal Model for Image Generation and Editing}

\author{%
  Yuxin Song$^{1}\textsuperscript{*}$, 
  Wenkai Dong$^{1}\textsuperscript{*}$, 
  Shizun Wang$^{2}\textsuperscript{*}$,
  Qi Zhang$^1$,
  Song Xue$^1$,\\
  \textbf{Tao Yuan$^1$, Hu Yang$^1$, Haocheng Feng$^1$, Hang Zhou$^1$, Xinyan Xiao$^1$, Jingdong Wang$^1\textsuperscript{\Envelope}$} \\
  $^1$Baidu VIS \quad 
  $^2$National University of Singapore \quad
  \\
  {\small $^{*}$ Equal Contribution \qquad \Envelope~Corresponding Author} \\
}

\begin{document}

\maketitle

\begin{abstract}

Unified Multimodal Models (UMMs) have demonstrated remarkable performance in text-to-image generation (T2I) and editing (TI2I), whether instantiated as assembled unified frameworks which couple powerful vision-language model (VLM) with diffusion-based generator, or as naive Unified Multimodal Models with an early fusion of understanding and generation modalities.
We contend that in current unified frameworks, the crucial capability of multimodal generative reasoning which encompasses instruction understanding, grounding, and image referring for identity preservation and faithful reconstruction, is intrinsically entangled with high-fidelity synthesis.
In this work, we introduce {\ours}, a novel approach that bridges the VLM and diffusion model via {a multimodal \textit{``kontext''}} composed of semantic cues and coarse-grained image conditions encoded from multimodal inputs. 
This design delegates the complex ability of multimodal generative reasoning to powerful VLM while reserving diffusion model’s role for high-quality visual synthesis.
To achieve this, we propose {a three-stage progressive training strategy}. First, we connect the VLM to a lightweight diffusion head via multimodal kontext tokens to {unleash} the VLM’s generative reasoning ability. 
Second, we {scale} this head to a large, pre-trained diffusion model to enhance visual detail and realism.
{Finally, we introduce a low-level image encoder to improve image fidelity and perform instruction tuning on downstream tasks.}
Furthermore, we build a comprehensive data pipeline integrating real, synthetic, and 
open-source datasets, covering diverse multimodal reference-to-image scenarios, including image generation, instruction-driven editing, customized generation, and multi-subject composition. Experiments show that our approach matches strong unified baselines and even outperforms task-specific state-of-the-art methods in several cases.






\begin{figure}[t!]
\centering
\includegraphics[width=1.0\linewidth]{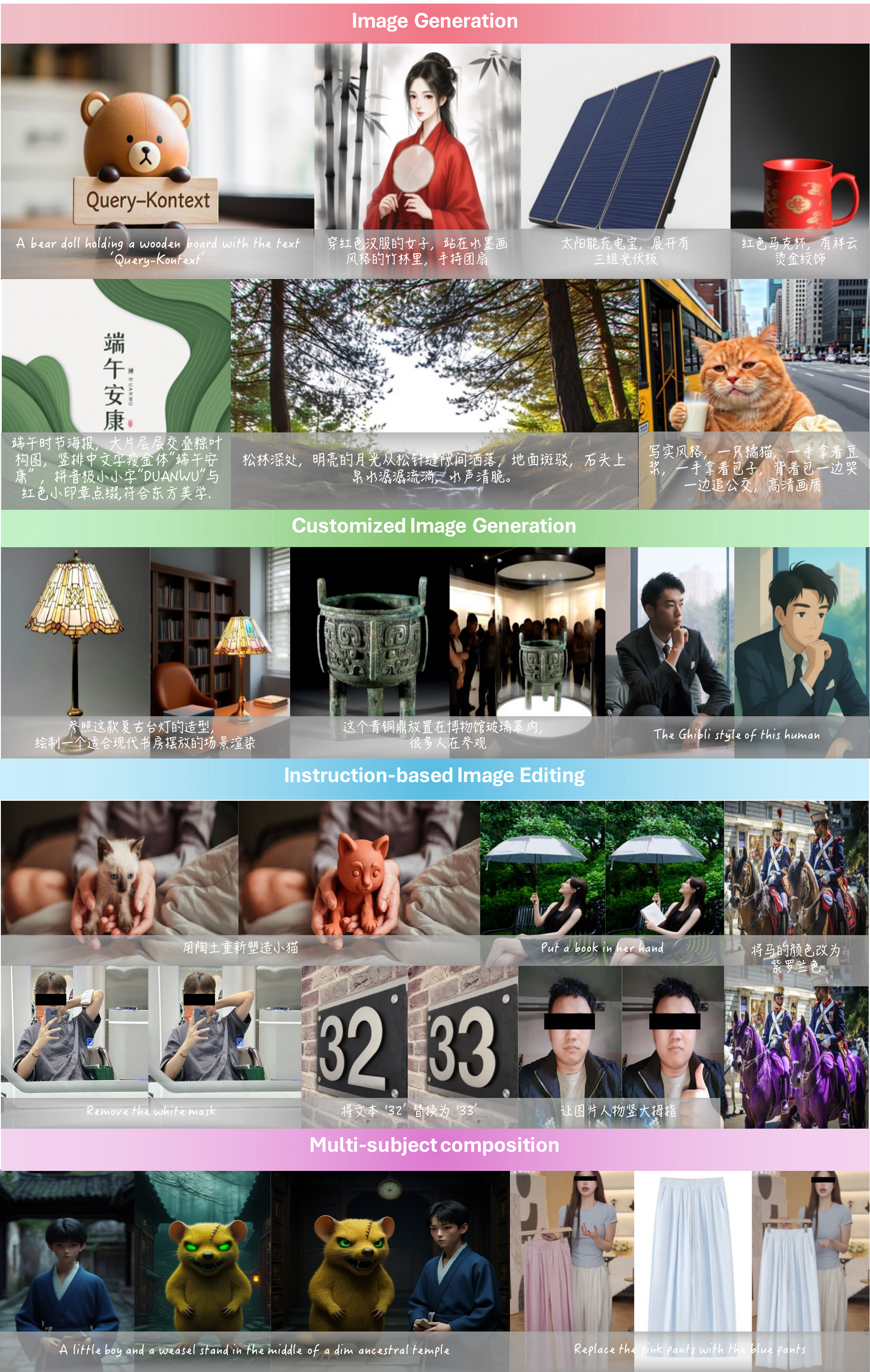}
\vspace{-15pt}
\caption{Showcase of Query-Kontext model on multimodal reference-to-image tasks.}
\vspace{-15pt}
\label{fig:cover_img}
\end{figure}

\end{abstract}    
\section{Introduction}

Unified Multimodal Models (UMMs) have recently achieved notable progress in both image generation (T2I) \cite{sdxl, SD3, chen2024pixart, ho2022cascaded, flux2024, deng2025bagel, gao2025seedream, liu2025step1x, metaquery, g4o_image, wu2025qwenimagetechnicalreport} and editing (TI2I) \cite{brooks2023instructpix2pix, zhang2023magicbrush, ye2025imgedit, liu2025step1xeditpracticalframeworkgeneral, labs2025flux_kontext, googleGemini2, kuprashevich2025nhr_edit, wang2025seededit, wei2024omniedit}.
Two prominent design paradigms have emerged from this work. The first assembled unified framework leverages external diffusion transformers, such as MMDiT \cite{SD3, dit}, which are paired with off-the-shelf vision–language models (VLMs) or large language models (LLMs) to provide semantic conditioning.
The second paradigm, naive UMMs, integrate generation and understanding more tightly through mixed-modal early-fusion transformers \cite{transfusion, deng2025bagel, chameleon, januspro2025, emu3, tong2024metamorph, janusflow2024}, where autoregressive modules with strong reasoning ability are jointly trained with diffusion modules specialized in visual synthesis.

While these paradigms expand task coverage and streamline deployment, they also entangle multimodal generative reasoning and high-fidelity rendering. Consequently, the unique strengths of VLMs (semantic understanding, grounding, structured reasoning \cite{Qwen2-VL, qwenvl, qwen2.5-vl, internvl2.5, internvl2, yao2024dense, zhao2024octopus, yao2024mulberry}) and diffusion models (photorealistic synthesis and detail fidelity \cite{dit, chen2023pixart, imddpm, ddpm, chen2024pixartalpha, chen2024pixartsigma, ho2022cascaded}) cannot be fully exploited. 
We identify two sources of this limitation. First, \textit{assembled} unified frameworks typically use a frozen VLM or LLM as a static feature extractor, narrowing the conditioning signal to only high-level semantics for the diffusion generator.
Second, \textit{native} UMMs force generative reasoning and visual rendering to be optimized jointly, introducing capacity competition and hindering generalization, particularly when tasks demand both fine-grained edits and strong semantic control.
While attempts to mitigate these issues through methods like mixture-of-experts (e.g., LlamaFusion \cite{shi2024llamafusion}) or mixture-of-transformers (e.g., BAGEL \cite{deng2025bagel}) have been made, they only partially alleviate the tension.

In this work, we propose \textbf{\ours}, an economic ensemble UMM that leverages the multimodal \textit{``kontext''}  composed of semantic and coarse image conditions to cleanly decouple the generative reasoning of VLM from the high-fidelity rendering of diffusion model.
To realize this separation, we develop a three-stage progressive training strategy.
\textbf{Stage 1:} Bridge the VLM to a lightweight diffusion head through \textit{``kontext''} tokens.
Using parameter-efficient fine-tuning (LoRA) \cite{hu2021lora}, we \textbf{unleash} the potential of VLM and steer it toward multimodal generative reasoning skills such as instruction following, spatial grounding, and identity-preserving image referencing. 
\textbf{Stage 2:} \textbf{Scale} the lightweight head to a well-trained large diffusion model (roughly 10$\times$ more parameters). We re-align both the text and \textit{``kontext''} tokens from the VLM to the scaled diffusion model by using text-to-image generation and image-reconstruction objectives.
\textbf{Stage 3:} Introduce a low-level image encoder \cite{SD-vae} that injects fine-grained structural and textural cues into the diffusion model while keeping the VLM frozen. This step strengthens identity preservation \cite{ye2023ipadaptor, wang2024instantid, wu2025uno_flux, song2025insertanything} and reconstruction fidelity in \cite{liu2025step1xeditpracticalframeworkgeneral, ye2025imgedit, huang2024smartedit, xu2025insightedit} challenging editing scenarios.

In summary, our contributions are:
\begin{itemize}
    \item We propose \textbf{\ours}, an  economic ensemble UMM that decouples multimodal generative reasoning in VLMs from the high-fidelity visual rendering performed by diffusion models.
    \item We present a three-stage progressive training strategy that progressively aligns the VLM with increasingly capable diffusion generators while amplifying their respective strengths in generative reasoning and visual synthesis.
    \item We present a deliberate dataset curation scheme to collect real, synthetic, and carefully filtered open-source datasets to cover diverse multimodal reference-to-image scenarios.
\end{itemize}

\section{\ours}
\noindent In this work, we propose {\ours}, a unified multimodal model for image generation and editing that delegates multimodal generative reasoning to the VLM while reserving the diffusion model's capability for high-quality visual synthesis. In Sec \ref{sec:architecture}, we present the architectural design of the {\ours} model (Figure \ref{fig:model}). In Sec \ref{sec:training}, we design a three-stage progressive learning strategy and introduce the details of training recipe (Figure \ref{fig:training}). In Sec \ref{sec:implementation}, we introduce the implementation details of model hyper-parameters and infrastructures.

\subsection{Architecture} \label{sec:architecture}
\begin{figure}[t!]
\centering
\includegraphics[width=1.0\linewidth]{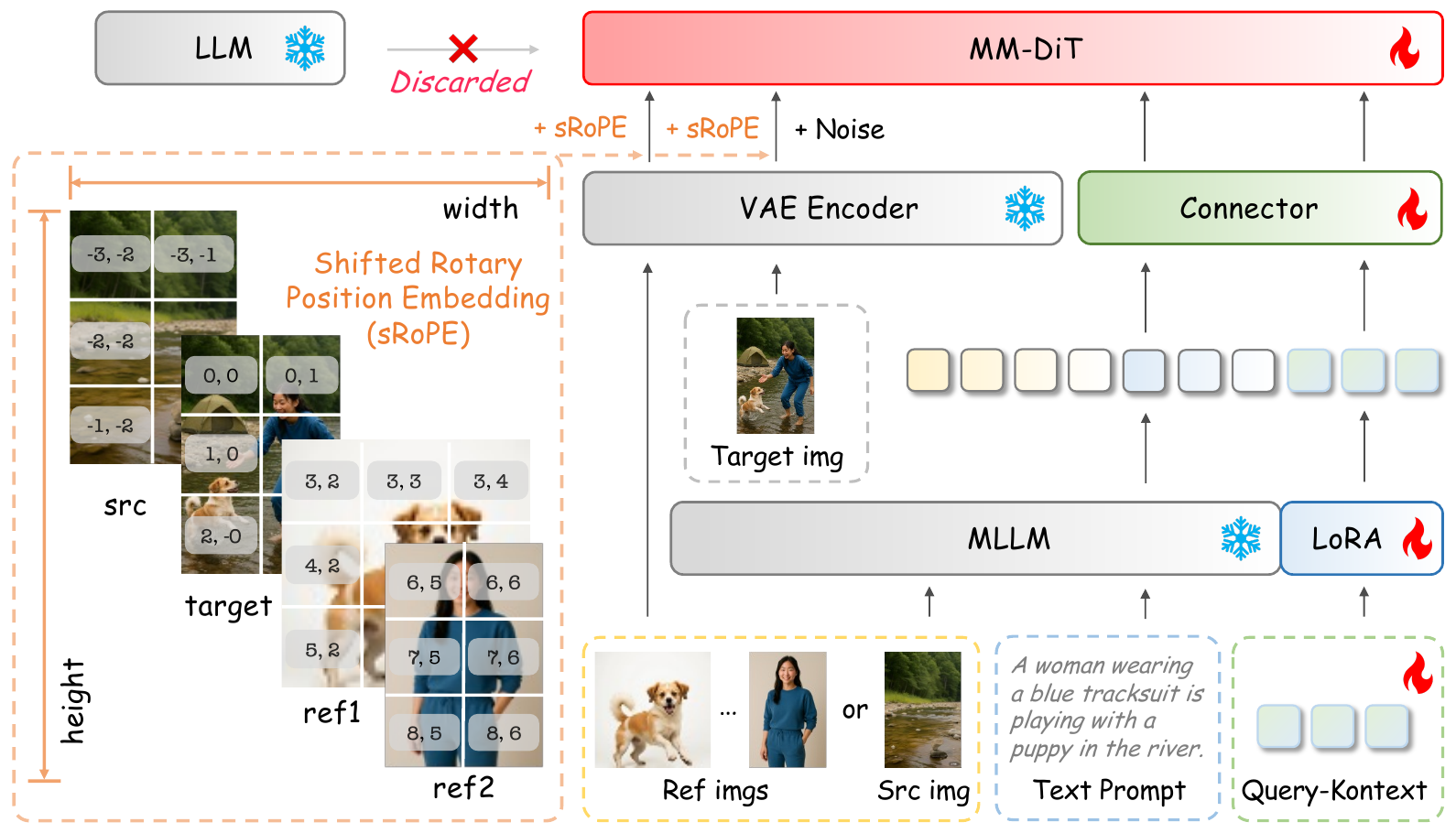}
\caption{\textbf{The overall framework of the unified multi-modal to image generation and editing model, {\ours}.} 
}
\label{fig:model}
\end{figure}

As shown in Figure \ref{fig:model}, {\ours} comprises four main components: a Multimodal Large Language Model (MLLM), a connector module, a Multimodal Diffusion Transformer (MMDiT), and a low-level image encoder (VAE). The MLLM is initialized with the Qwen2.5-VL model \cite{qwen2.5-vl}, which encodes and fuses multimodal inputs including the text prompt, input image(s), and a set of learnable query tokens.
The output is a fixed-length sequence of $kontext$ tokens $Q = \{q_1,\dots,q_K\}$ which serves as coarse image-level conditioning for the diffusion decoder while providing high-level semantic cues. 
Intuitively, the $kontext$ tokens $Q$ encode what content should appear in the output image (the semantic information from the text prompt) and how the output should incorporate visual cues from the provided input images, as enforced by the training supervision in Sec.~\ref{sec:training_s1}. 
The $kontext$ $Q$ and text tokens $T$ are passed through a lightweight connector module to align them with the diffusion model’s latent space. In practice, we concatenate the connector-generated text embeddings with the kontext embeddings, thereby enriching the semantic context available to the diffusion model.

We initialize the diffusion model using our in-house MMDiT model 
and replace its original text encoder with the MLLM (\textit{training details for this alignment are discussed in Sec.~\ref{sec:training}}).
We concatenate the sequence of text $T$ and  $kontext$ tokens $Q$ from the MLLM with: (i) the noisy image latent at the current diffusion step $t$, and (ii) the low-level visual feature tokens extracted from the input image(s) by the VAE. The concatenated sequence is then fed into the MMDiT model in an in-context manner \cite{labs2025flux_kontext, zhang2025ICEdit}, allowing the diffusion model to attend to both the textual prompt and the visual cues from the input images.

Moreover, we distinguish between naive UMMs and assembled UMMs in Table \ref{tab:position}. The comparison highlights whether each model trains from scratch, freezes parameters, or adapts pretrained components, and specifies the flow of information through text embeddings (TE), low-level image embeddings (LE), and query embeddings (QE). 
In particular, query embeddings naturally unleash the in-context learning capabilities of the VLM, enabling the model to reason over multimodal inputs and generate coherent images.
Unlike prior methods, \textbf{our {\ours} integrates query embeddings alongside text and low-level image embeddings, while effectively decoupling understanding and generation modules for improved efficiency and flexibility.}

\begin{table}[htbp]
\centering
    \begin{tabular}{l|ccc|ccc}
    \toprule
    \multirow{2}{*}{\textbf{Method}} & \multicolumn{3}{c|}{\textbf{Module}} & \multicolumn{3}{c}{\textbf{Information}} \\
    \cmidrule(lr){2-4}\cmidrule(lr){5-7}
    & \textbf{Understanding} & \textbf{Connector} & \textbf{Generation} & \textbf{TE} & \textbf{LE} & \textbf{QE} \\
    \midrule
    \rowcolor{gray!20} \multicolumn{1}{l}{\textsl{Native UMMs}}  & \multicolumn{6}{l}{} \\
    Janus-Pro \cite{januspro2025}&  \includegraphics[width=0.35cm]{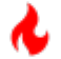} & - & \includegraphics[width=0.35cm]{figs/fire.pdf} & \cmark & \xmark & \xmark \\
    OmniGen2 \cite{wu2025omnigen2}&  \includegraphics[width=0.35cm]{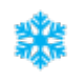} \rarrow \includegraphics[width=0.35cm]{figs/fire.pdf} & - & \includegraphics[width=0.35cm]{figs/fire.pdf} & \cmark & \cmark & \xmark \\
    BAGEL \cite{deng2025bagel}&  \includegraphics[width=0.35cm]{figs/fire.pdf} & - & \includegraphics[width=0.35cm]{figs/fire.pdf} & \cmark & \xmark & \xmark \\
    \midrule
    \rowcolor{gray!20} \multicolumn{1}{l}{\textsl{Assembled UMMs}}  & \multicolumn{6}{l}{} \\
    
    Metaquery \cite{metaquery}& \includegraphics[width=0.35cm]{figs/froze.pdf} & \includegraphics[width=0.35cm]{figs/fire.pdf} &  \includegraphics[width=0.35cm]{figs/froze.pdf} \rarrow \includegraphics[width=0.35cm]{figs/fire.pdf} & \cmark & \xmark & \cmark \\
    Step1X-Edit \cite{liu2025step1xeditpracticalframeworkgeneral}& \includegraphics[width=0.35cm]{figs/froze.pdf} & \includegraphics[width=0.35cm]{figs/fire.pdf} &  \includegraphics[width=0.35cm]{figs/froze.pdf} \rarrow \includegraphics[width=0.35cm]{figs/fire.pdf} & \cmark & \cmark & \xmark \\
    Uniworld-v1 \cite{lin2025uniworldv1}& \includegraphics[width=0.35cm]{figs/froze.pdf} & \includegraphics[width=0.35cm]{figs/fire.pdf} &  \includegraphics[width=0.35cm]{figs/froze.pdf} \rarrow \includegraphics[width=0.35cm]{figs/fire.pdf} & \cmark & \cmark & \xmark \\
    FLUX.1 Kontext \cite{labs2025flux_kontext}& \includegraphics[width=0.35cm]{figs/froze.pdf} & - & \includegraphics[width=0.35cm]{figs/fire.pdf}  & \cmark & \cmark & \xmark \\
    Qwen-Image \cite{wu2025qwenimagetechnicalreport}&  \includegraphics[width=0.35cm]{figs/froze.pdf} & - & \includegraphics[width=0.35cm]{figs/fire.pdf} & \cmark & \cmark & \xmark \\
    
    \textbf{{\ours} (Ours)} & \includegraphics[width=0.35cm]{figs/froze.pdf} \rarrow \includegraphics[width=0.35cm]{figs/fire.pdf} & \includegraphics[width=0.35cm]{figs/fire.pdf} & \includegraphics[width=0.35cm]{figs/froze.pdf} \rarrow \includegraphics[width=0.35cm]{figs/fire.pdf} & \cmark & \cmark & \cmark \\
    \bottomrule
    \end{tabular}

\vspace{8pt}
\caption{
    \textbf{Comparison of mainstream unified multimodal models on the modeling paradigms and the information flow.} $\includegraphics[width=0.35cm]{figs/fire.pdf}$ denotes training from scratch, $\includegraphics[width=0.35cm]{figs/froze.pdf}$ indicates freezing the parameters during training and 
    $\includegraphics[width=0.35cm]{figs/froze.pdf} {\rightarrow} \includegraphics[width=0.35cm]{figs/fire.pdf}$
    represents training from a pretrained model. 
    For the input modalities, ``TE'' refers to text embeddings, ``LE'' to low-level image embeddings, and ``QE'' to query embeddings.
}
\label{tab:position}
\end{table}

Furthermore, we design a \textit{shifted 2D Rotary Position Embedding} (RoPE) scheme \cite{su2024roformer, wu2025uno_flux} to incorporate multi-image positional conditioning and avoid confusion among multiple reference images (as illustrated in Figure \ref{fig:model}). 
In the standard diffusion architecture, each spatial position of a latent feature map (with size of $h \times w$) is identified by a 2D index $(i, j)$, where $i \in [0,\,w-1]$ and $j \in [0,\,h-1]$. 
We introduce a task-specific prior to adjust these coordinates based on the fidelity requirements of the input images. 
For tasks requiring pixel-level fidelity to an input image (e.g., instruction-based editing), we treat the input image as a \textit{source image}, denoted $img_{src}$. 
For tasks requiring identity preservation (e.g., personalized generation or multi-image composition), we treat the input image as a \textit{reference image}, denoted $img_{ref}$. 
We then shift the coordinate indices of the VAE latent for each image type accordingly: for reference image latents, we shift indices into the positive quadrant, whereas for the source image latent, we shift into the negative quadrant.
we define the coordinates for the $n$-th reference latent as: 
\begin{equation}
    \big(i_\text{ref}^n,\, j_\text{ref}^n\big) = \big(i + w*n,  j + h*n\big)
\end{equation}
where $i \in [0, w-1]$, $j \in [0, h-1]$ and $n \in [1,N]$. 
Meanwhile, for the source image latent 
we shift the coordinates in the negative direction:
\begin{equation}
(i'_{\text{src}},\, j'_{\text{src}}) \;=\; (-\,i, \;\; -\,j)\,,
\end{equation}
where $i \in [0, w-1]$, $j \in [0, h-1]$ and $n \in [1,N]$. 
Finally, we add the shifted RoPE on the feature maps of the input image latent(s) and the noisy latent at their respective shifted coordinates (i.e., added element-wise to each spatial location).

\subsection{Individualized-Teaching Curriculum}\label{sec:training}
\begin{figure}[t!]
\centering
\includegraphics[width=1.0\linewidth]{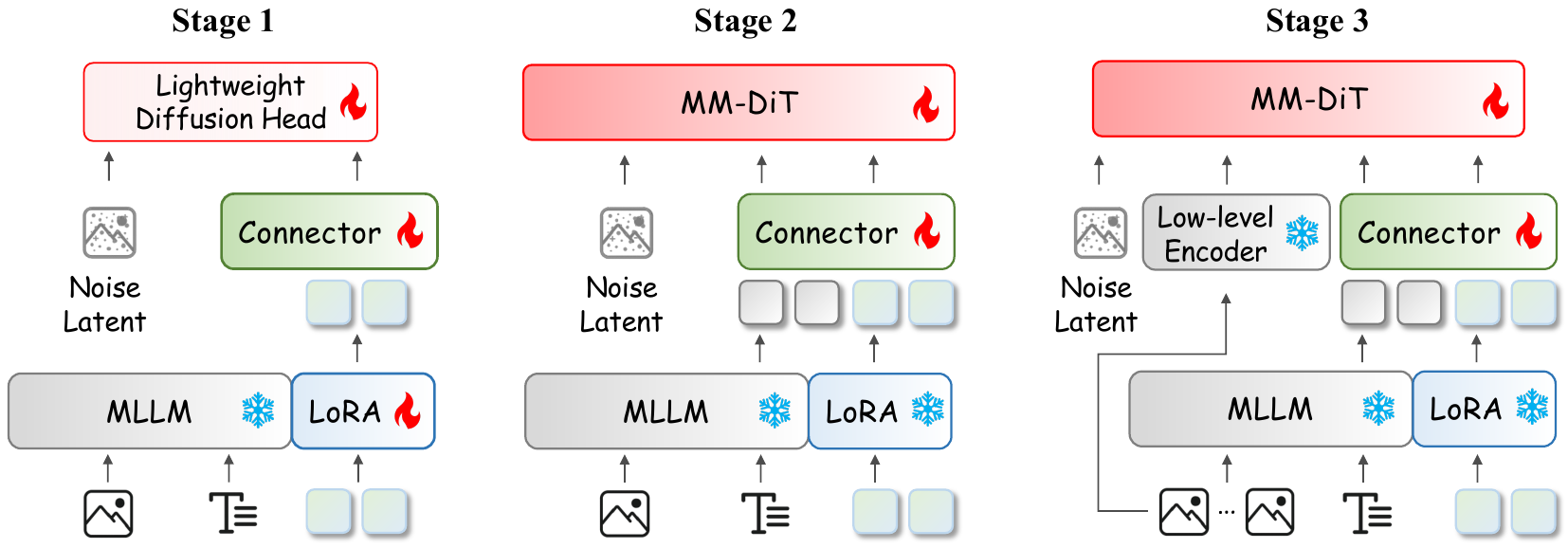}
\caption{\textbf{Three training stages of {\ours}.} Note that the Diffusion Head is only used in the Stage 1. In the Stage 2 and 3, we scale up Diffusion model to 10× parameters and keep MLLM frozen to provide coarse-grained image conditions.
}
 \vspace{-10pt}
\label{fig:training}
\end{figure}

As shown in Figure \ref{fig:training}, we propose a three-stage progressive training strategy that both unlocks the generative reasoning capabilities of the VLM and progressively aligns it with increasingly powerful diffusion generators. As a result, {\ours}, guided by multimodal $kontext$ tokens, effectively decouples the multimodal generative reasoning of the VLM from the high-fidelity visual rendering carried out by diffusion models.

\noindent\textbf{Stage 1:} 
We unleash the generative reasoning potential of the MLLM through two key architectural designs: we first use learnable query tokens (``kontext'') to represent a mixture of semantic cues and coarse-grained image conditions, and then align the output $kontext$ tokens with a lightweight diffusion head that performs noisy prediction at a coarse level. 
We train all parameters of the connector, the diffusion head, and the MLLM's LoRA modules on a trio of tasks: text-to-image generation, image reconstruction, and image transformation (see Section ~\ref{sec:data_cluster}). This training methodology preserves the MLLM's inherent language-vision understanding while cultivating its emergent ability for multimodal generative reasoning.
\label{sec:training_s1}

\noindent\textbf{Stage 2:} 
Next, we replace the lightweight diffusion head with our in-house diffusion model based on the MMDiT architecture for high-fidelity generation. In Stage 2, the full MLLM parameters (the LoRA parameters are merged into the MLLM) remain frozen, and we optimize the $kontext$ tokens, the connector, and all parameters of the large diffusion model. 
In preliminary experiments, we observed that completely freezing the diffusion model was feasible for smaller head but failed for a larger diffusion model (\textit{the experiments details and discussion are available in Section \ref{sec:disscusion}}). Therefore, we allow the diffusion model to be full-parameters fine-tuning in this stage. 
To keep training efficient, {\ours} is trained only on text-to-image generation and image reconstruction tasks in this stage, which accelerates convergence and reduces training cost for fast alignment from MLLM to the diffusion model.

\noindent\textbf{Stage 3:}
Finally, we introduce a dedicated low-level image encoder for source or reference images to further refine the diffusion model for high-fidelity image referring. In Stage 3, the MLLM remains fully frozen, and we optimize only the {\ours} tokens and the connector. Additionally, we apply the LoRA-based fine-tuning to the diffusion model itself to preserve its high-quality image synthesis ability while extending it to all our tasks. This includes not only standard text-to-image generation but also instruction-guided image editing, user-customized image generation, and multi-subject composition tasks.

\subsection{Implementation}\label{sec:implementation}
\noindent \textbf{Architecture.} We initialize the MLLM from Qwen2.5-VL-7B and implement the connector as a two-layer MLP. (\textit{details of architecture configuration are provided in Table \ref{tab:arch}.})
The connector maps text and kontext tokens into the diffusion latent space; the outputs are concatenated before being fed to the diffusion transformer.
Moreover, we implement the diffusion head in the stage a with a lightweight MMDiT architecture (${\sim}$870$M$ parameters). 
We set the max reference images $N=2$ and {$K=128$} in Query-kontext $Q = \{q_1,\dots,q_K\}$. 
We set {rank $r_d=256, \alpha_d=256$ in the diffusion model's and rank $r_m=128, \alpha_m=256$ in the MLLM's LoRA.}

\begin{table}[t!]
    \centering
    \caption{Configuration of {\ours} architecture.}
    \label{tab:arch}
    \begin{tabular}{l|cc|cc|c|c}
    \toprule
    \multirow{2}{*}{\textbf{Configuration}}
      & \multicolumn{2}{c|}{\textbf{MLLM}} 
      & \multicolumn{2}{c|}{\textbf{VAE}} 
      & \multirow{2}{*}{\textbf{Connector}}
      & \multirow{2}{*}{\textbf{MMDiT}} \\
    \cmidrule{2-5}
      & ViT & LLM & Enc & Dec & & \\
    \midrule
    \# Layers & 32 & 28 & 8 & 14 & 2 & 42 \\
    \# Num Heads~(Q / KV) & 16 / 16 & 28 / 4 & - & - & - & 40 / 40 \\
    Head Size & 80 & 128 & - & - & - & 64 \\
    Intermediate Size & 3,456 & 18,944 & - & - & - & 10240 \\
    Patch / Scale Factor & 14 & - & 8x8 & 8x8 & - & 2 \\
    Channel Size & - & - & 16 & 16 & - & - \\
    \midrule
    \# Parameters & \multicolumn{2}{c|}{7B} & 34M & 50M & 5.9M & 10B \\
    \bottomrule
    \end{tabular}
\end{table}

\noindent \textbf{Training recipe.}
The default configuration on the resolution with $512\times512$ is provided in Table \ref{tab:recipe}. After Stage 3, we introduce a resolution upscaling stage using the same mixed multi-task dataset at a higher resolution. In this stage, the training resolution is increased to $1024\times1024$, the learning rate is further reduced to $1\times 10^{-5}$, and training continues for an additional 3,000 steps with a global batch size of 256.

\noindent \textbf{Infrastructure.}
We adopt a hybrid parallel optimization strategy during training.
we enable tensor parallelism on the VLM side. For the diffusion model, we use parameter sharding (ZeRO Stage-2) together with bfloat16 (BF16) mixed-precision training. To keep sequence lengths uniform within a mini-batch, we maintain two independent bucketeers—by image aspect ratio (supporting 1:1, 1:2, 2:3, 3:4, 3:5, 4:5, and 9:16) and by the number of reference images—so that samples in the same batch produce the same number of latent tokens, reducing padding and improving throughput.

\begin{table}[t]
\caption{
    \textbf{The data outline and training details about each training stage.}  Where, $Q.$ denotes the Query-kontext tokens, $Con.$ is Connector module. 
    }
\label{tab:recipe}
\centering
\scalebox{1.0}{
    \begin{tabular}{l|c|c|c}
    \toprule
    {\textbf{Stage}} & \textbf{Stage 1} & \textbf{Stage 2} & \textbf{Stage 3} \\
    \midrule
    \multirow{3}*{Task} & Image Generation & Image Generation  & Instruction Editing \\
     & Image Reconstruction  & \multirow{2}*{Image Reconstruction} & Customized Generation \\
     &  Image Transformation &  &  Multi-subject \\
     \midrule
    Type & {T2I, I2I, TI2I} & {T2I, I2I} & {T2I, TI2I} \\
     \midrule
    \multirow{2}*{Training Param.} & MLLM's LoRA, $Con.$, & \multirow{2}*{$Con.$, MMDiT, $Q.$} & MMDiT's LoRA,  \\
     & Diffusion head, $Q.$ &  & $Con.$, $Q.$ \\
     \midrule
    Global Batch Size & 512 & 1024 & 512 \\
    Steps (K) & 72 & 420 & 30 \\
    Learning Rate & 1e-4 & 1e-4 & 2e-5 \\
    \bottomrule
    \end{tabular}
}
\end{table}
\section{Data Curation}

\begin{table}[h]
    \centering
    \caption{\textbf{The data outline about each training stage.} $^{\dagger}$ denotes only the Chinese prompt.}
    \label{tab:data_overview}
    \begin{tabular}{c|c|c|c}
    \toprule
    \textbf{Stage} & \textbf{Task} & \textbf{Data source} & \textbf{Size} \\ 
    \midrule
    \multirow{2}{*}{1, 2, 3}
    & Image generation, & ShareGPT-4o-Image\citep{chen2025sharegpt}, BLIP3o\citep{chen2025blip3} & 30M \\ 
    & Image reconstruction & in-house real data$^{\dagger}$ & 170M \\ 
    \midrule
    {1} & Image transformation & mmc4\citep{zhu2023mmc4}, OmniCorpus\citep{li2024omnicorpus} & 800K \\ 
    \midrule
    \multirow{7}{*}{3} & \multirow{3}{*}{Instruction editing} & NHR-Edit\citep{kuprashevich2025nhr_edit}, GPT-Edit\citep{wu2025qwenimagetechnicalreport}, OmniEdit\citep{wei2024omniedit} & 3M \\ 
    &  & in-house video data & 2M \\ 
    &  & in-house real data & 300K \\ 
    \cmidrule{2-4}
    & \multirow{2}{*}{Customized generation}  & subject200k\citep{tan2024ominicontrol} & 200K \\ 
    & & in-house real data & 1.8M \\ 
    \cmidrule{2-4}
    
    & \multirow{2}{*}{Multi-subject composition} & MUSAR-Gen\citep{guo2025musar} & 29K \\ 
    & & G4o synthesis data & 40K \\ 
    \bottomrule 
    \end{tabular}
\end{table}

We constructed a multimodal reference-to-image dataset (as summarized in Table \ref{tab:data_overview}) comprising a mixture of real, synthetic, and carefully curated open-source datasets. This dataset spans five categories of tasks: text-to-image generation, image transformation, instruction editing, customized generation, and multi-subject composition.

\noindent \textbf{Text-to-Image Generation and Image Reconstruction.} We collected 30M open-source English image-text pairs (including ShareGPT-4o-Image \citep{chen2025sharegpt}, BLIP-3o \citep{chen2025blip3}, among others) as well as 170M in-house Chinese image-text pairs for text-to-image generation and image reconstruction tasks. The in-house data underwent extensive quality filtering based on image resolution, clarity, aesthetic score, watermark detection, and safety compliance. Among these Chinese data, 150M belong to general categories (balanced across diverse domains), and 20M come from specific vertical domains (e.g., artistic styles, logos, automobiles, text-containing images, celebrities, posters, etc.).

\begin{figure}[t]
\centering
\includegraphics[width=1.0\linewidth]{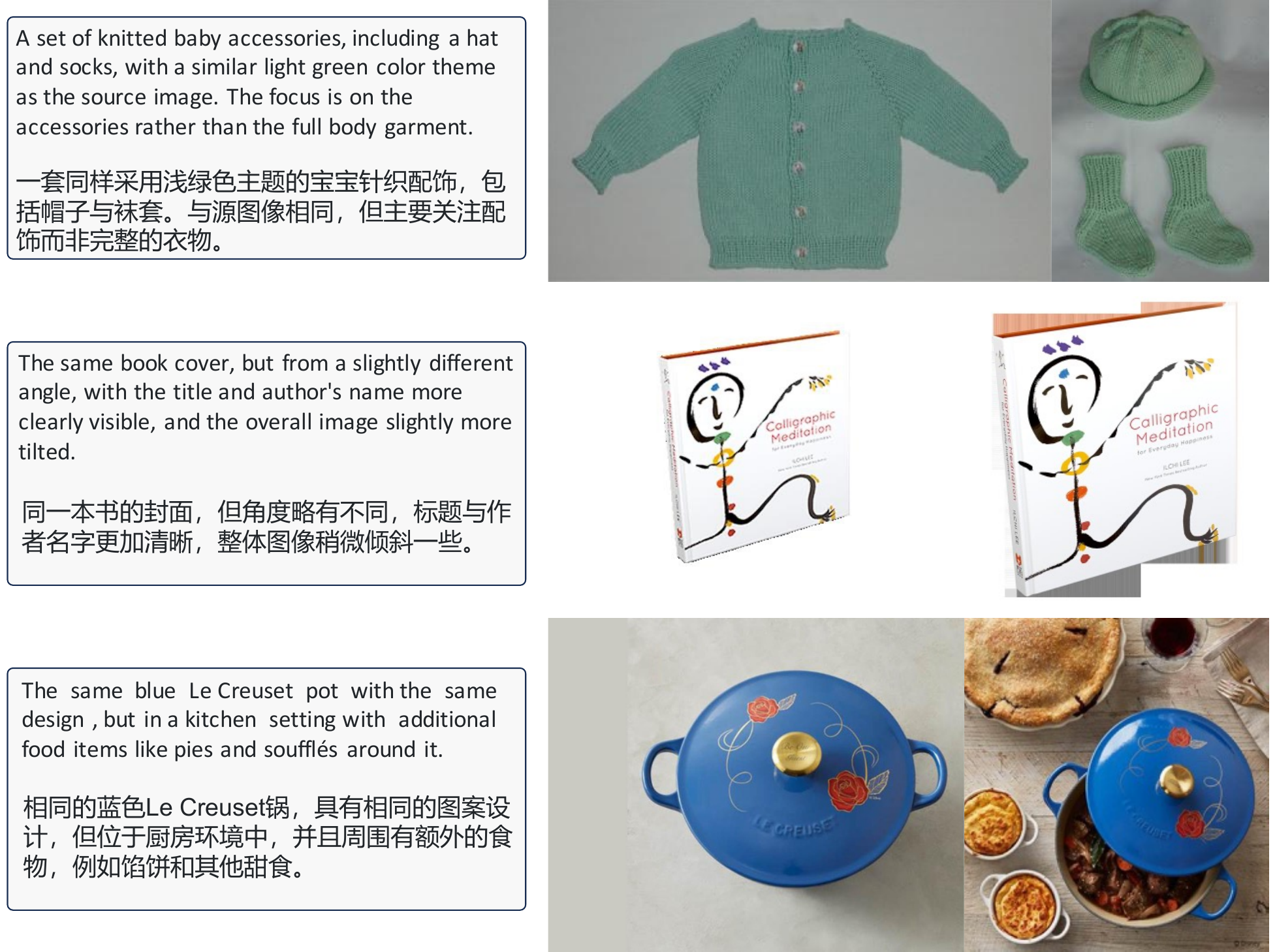}
\caption{Examples of the image transformation task. 
Each row shows a transformation instruction, the source image and the resulting target image, in order from left to right.}
\vspace{-10pt}
\label{fig:image_transformation_case}
\end{figure}

\noindent \textbf{Image Transformation.} \label{sec:data_cluster} Following the MetaQuery \citep{metaquery}, we constructed naturally occurring image pairs from web corpora \citep{chen2025multimodal, li2024omnicorpus} and generated corresponding open-ended transformation instructions by leveraging multi-modal large language models (MLLMs). Specifically, we clustered images that share the same accompanying caption from sources like MMC4-core\citep{chen2025multimodal}, OmniCorpus-CC \citep{li2024omnicorpus} and OmniCorpus-CW \citep{li2024omnicorpus} by using SigLIP \citep{tschannen2025siglip} image features, then filtered these clusters by a similarity threshold to obtain 0.8M image transformation triplets. 
As shown in Figure~\ref{fig:image_transformation_case}, each triplet contains a source image, an open-ended transformation instruction which is generated by Qwen2.5-VL as shown in Figure~\ref{fig:vlm_prompt}, and a target image. The instructions cover viewpoint changes (e.g., zoom-in/out, rotation), appearance modifications (e.g., color/material replacement), and structural adjustments (e.g., adding/removing objects). 

\begin{figure}[htbp]
\scriptsize
\begin{AcademicBox}[ Qwen2.5-VL Prompt]
The first image is the source image, the second image is the target image. create an interesting text 
prompt that can be used with the source images to generate the target image. 
\\ \\
This prompt should include:
- one general and unspecific similarity shared with the source image.
- all differences that only the target image has. 
This prompt should NOT include:
- any specific details that would allow generating the target image independently without referencing 
the source image. Remember the prompt should be concise and short (no more than 64 words).
\\ \\
The difference should include but not limited to:
Change in Angle. Describe the specific new angle, e.g.: 
* Side view, back view, viewed from the front side
* With a closer view, focus on the top of the item
* Cropped in the center, in a horizontal/vertical view
\\ \\
Same Subject, Altered Elements. Specify added/removed elements, e.g.:
* The same jacket but with a person wearing it
* Without the package
* with a bowl on the right
* Show the engine of the same car
\\ \\
Color Change. Describe the new color(s), e.g.:
* Blue and purple flowers instead of yellow
* Turn the color of the vase to cyan
* Same design but in white
\\ \\
Position Change. Specify the new position, e.g.:
* Put the item in the middle
* Move it to the side with a closer view
\\ \\
Background Change. Describe the modified background, e.g.:
* With a clean background
* The daylight turns dim
* Without a background
\\ \\
Quantity Change. State the updated quantity, e.g.:
* Show three trains of the same type
\\ \\
State/Process Change. Describe the transformation or action, e.g.:
* The beef is cooked
* The man is pouring batter into a pan
* Put it down
\\ \\
Here are some of the example:
1. Prompt: The complete front view of the same jersey top.
2. Prompt: Show three blue pot with same floral design, now placed in a cozy dining scene with food, 
drinks, and side dishes around it.
3. Prompt: The same pair of silver rings captured in another angle with brighter lighting and a clean 
white background and softer shadows.
\\ \\
Please generate one English prompt and one Chinese prompt, following the JSON format:
[`English prompt here', `Chinese prompt here'.]
\end{AcademicBox}
\caption{
   Example of the prompt used in Qwen2.5-VL to generate open-ended transformation instructions.
}
\label{fig:vlm_prompt}
\end{figure}

\noindent \textbf{Instruction Editing.} For the image editing instruction task, we first aggregated approximately 3M image-instruction-image triplets from open-source datasets, including NHR-Edit \citep{kuprashevich2025nhr_edit}(358k samples), GPT-Image-Edit \citep{gptimage}(1.5M samples), MagicBrush \citep{kawar2023imagic}(10k samples), and OmniEdit \citep{wei2024omniedit}(1.2M samples). We further filtered the MagicBrush subset using CLIP-based image and text similarity scores, and translated all datasets' instructions into Chinese using a large language model. 

\begin{figure}[t!]
\centering
\includegraphics[width=1.0\linewidth]{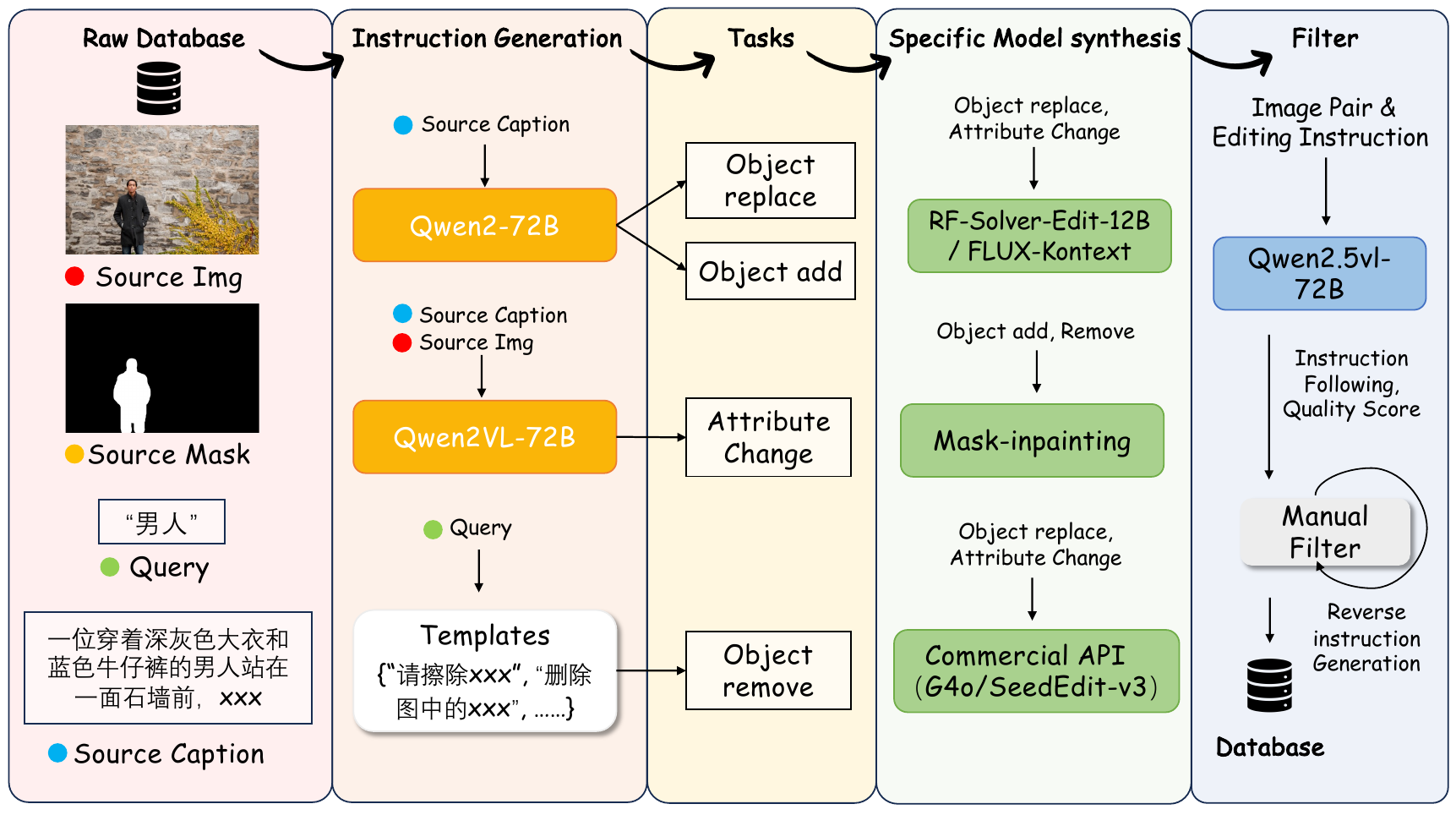}
\caption{ Examples of synthetic data pipeline for instruction Editing.
}
\label{fig:data_pipeline}
\end{figure}

\begin{figure}[t!]
\centering
\includegraphics[width=1.0\linewidth]{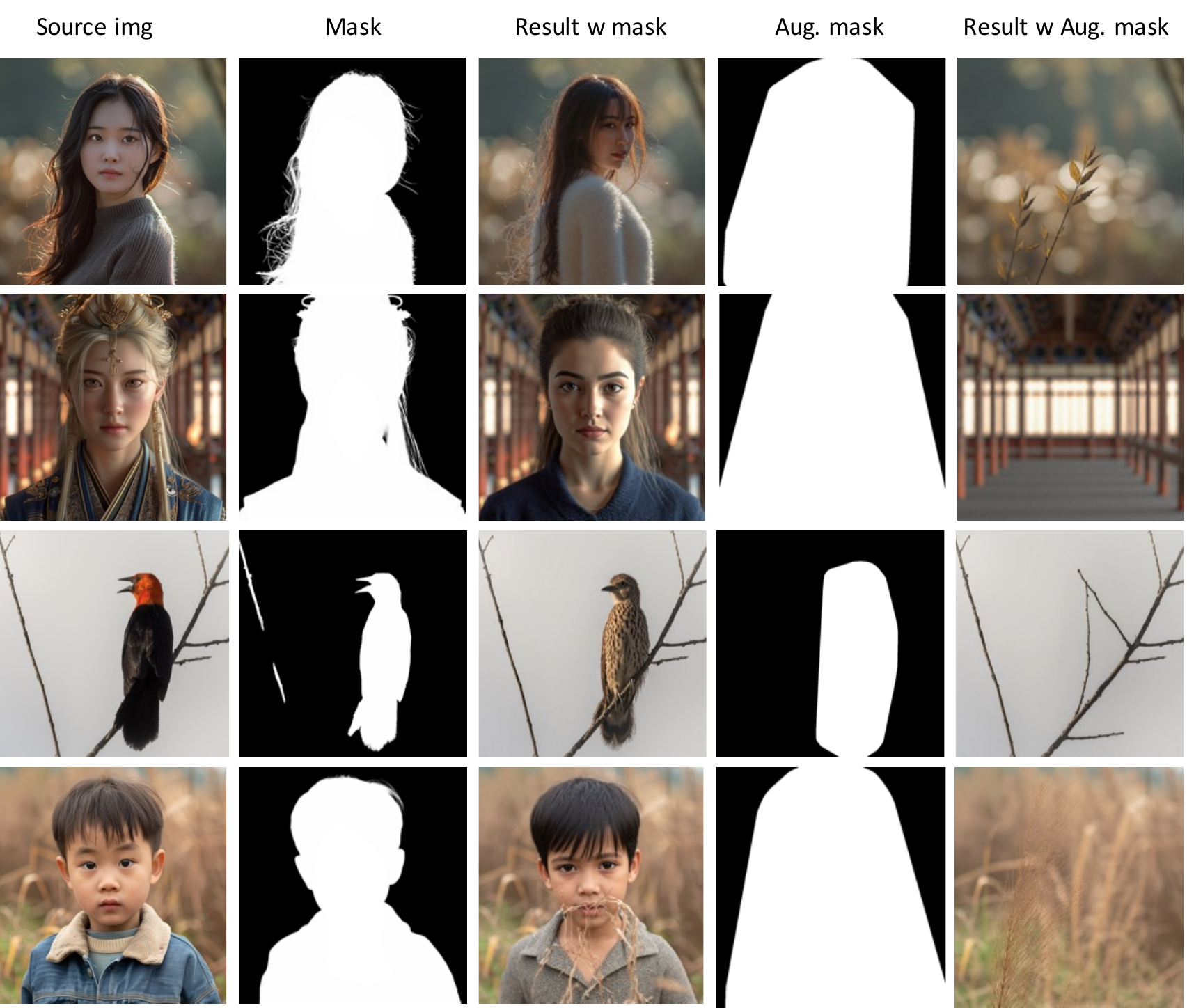}
\caption{ Examples of image inpainting with mask augmentation.
}
\label{fig:mask_aug_case}
\end{figure}

Building upon the methodologies of \citep{kuprashevich2025nhr_edit, wei2024omniedit, liu2025step1xeditpracticalframeworkgeneral}, we then constructed a synthetic data pipeline tailored for native Chinese instruction editing, producing an additional 300k high-quality triplets. 
As illustrated in Figure~\ref{fig:data_pipeline}. Given a source image, its segmentation mask, and a caption, we first extract object-level queries (e.g., “man”) and generate diverse editing instructions. Specifically, large language models (Qwen2-72B) generate textual editing prompts from captions, while multimodal models (Qwen2VL-72B) leverage both captions and images to produce more fine-grained attribute modification instructions. In addition, we incorporate template-based instructions (e.g., “remove xxx from the image”) to further handle object remove task. The generated instructions are categorized into four task types: object replacement, object addition, object removal, and attribute modification. Each task is then handled by specialized synthesis models: RF-Solver-Edit-12B \cite{wang2024rf_solver} or FLUX-Kontext \cite{labs2025flux_kontext} for replacement and attribute edits, mask-based inpainting model \cite{flux2024} for addition and removal, and commercial APIs (e.g., G4o/SeedEdit-v3) for more complex operations. Finally, the generated triplets are filtered through an automatic evaluation stage using Qwen2.5VL-72B, which scores instruction fidelity and image quality, followed by manual verification to ensure reliability. Finally, manual reverse instruction generation is applied by treating the source image as the target, ensuring supervision from authentic images without model-induced artifacts. Moreover, when applying mask-inpainting models to remove large objects, we adopt a mask augmentation strategy to mitigate the influence of shape-guided masks. Figure \ref{fig:mask_aug_case} presents a comparison between results with and without mask augmentation.

\begin{figure}[t!]
\centering
\includegraphics[width=1.0\linewidth]{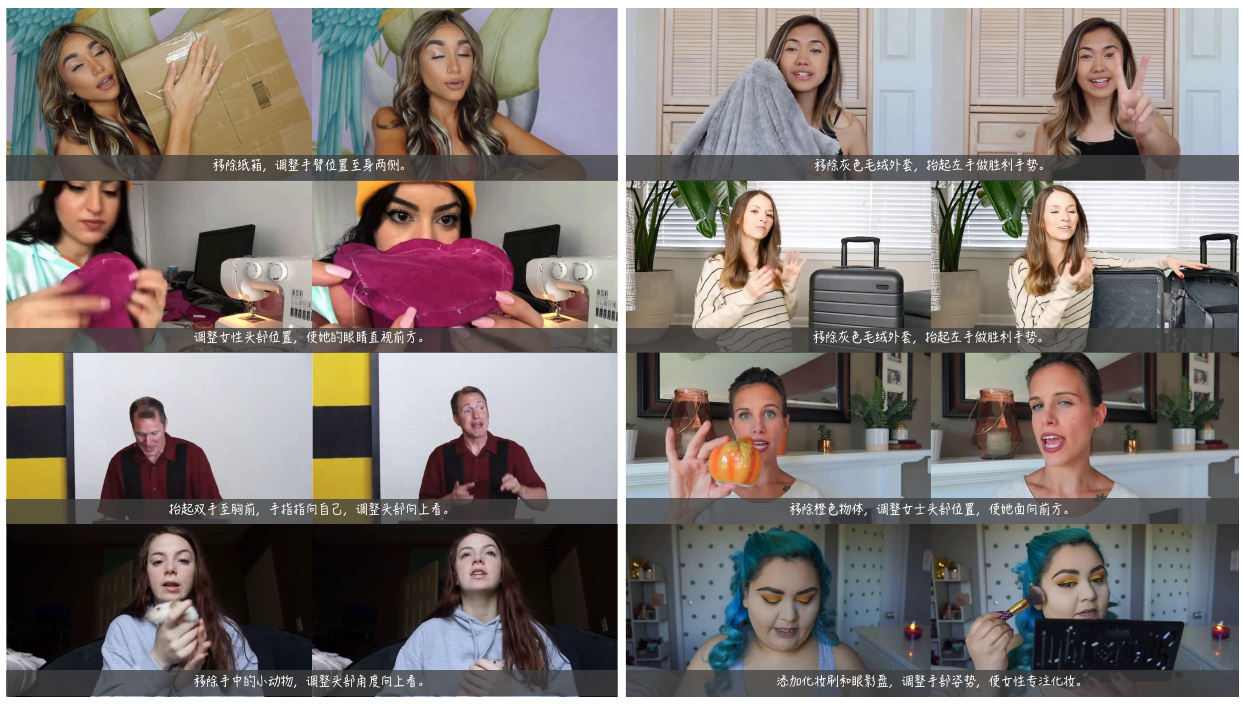}
\caption{ Examples of instruction editing data pair constructed from real video.
}
\label{fig:video_data_case}
\end{figure}

Finally, inspired by UniReal \citep{chen2025unireal}, we extended our dataset with video-based clusters derived from raw videos to cover more non-rigid editing tasks (e.g., motion changes, viewpoint shifts, view transitions such as zoom-in and zoom-out. 
Representative data examples are provided in Figure \ref{fig:video_data_case}.

\noindent \textbf{Customized Generation.} We leveraged the open-source Subject-200K \citep{tan2024ominicontrol} and UNO-1M \citep{wu2025uno_flux} datasets for customized (subject-driven) image generation. In addition, we augmented our data with portrait reference triplets synthesized using a dedicated model \footnote{\url{https://console.bce.baidu.com/qianfan/modelcenter/model/buildIn/detail/am-t3uhhjzbys6w}}
, which generates reference images of specific individuals. Through this approach, we accumulated approximately 0.3M portrait reference samples that maintain high facial similarity to the source subjects while exhibiting substantial diversity in poses, attire, and other attributes.

\begin{figure}[t!]
\centering
\includegraphics[width=1.0\linewidth]{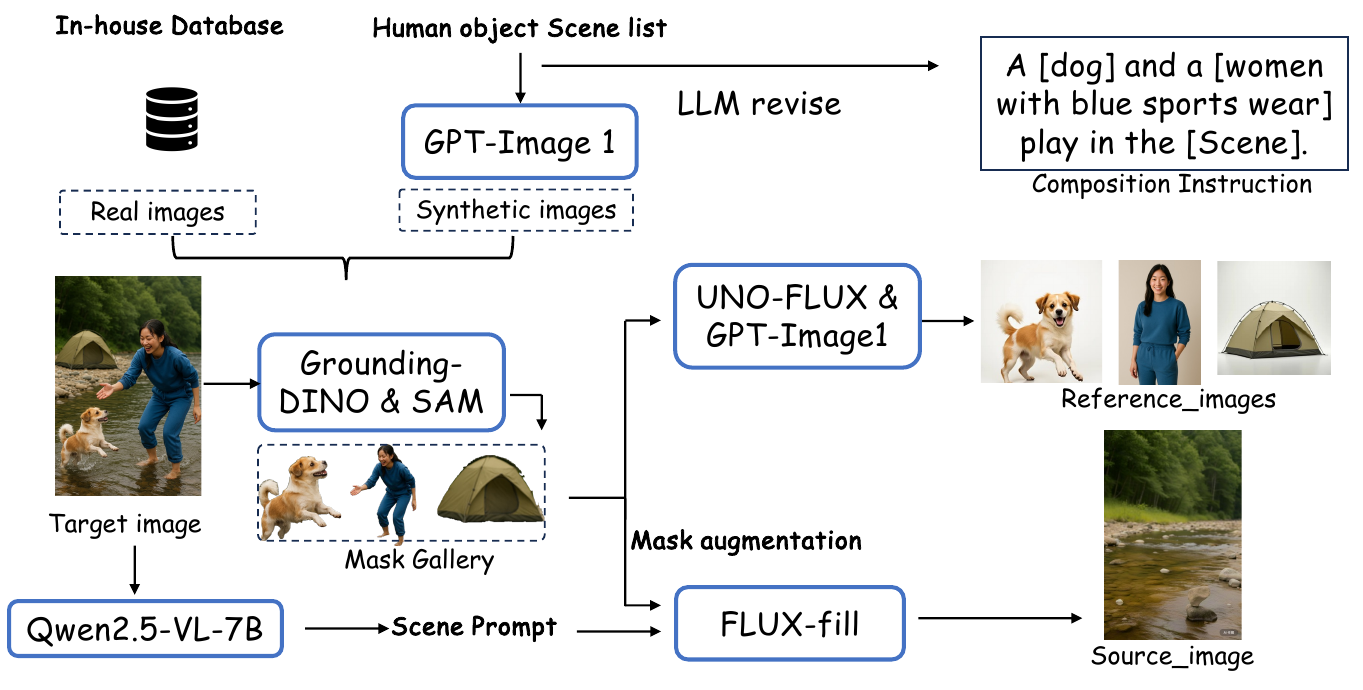}
\caption{ Examples of synthetic data pipeline for Multi-Subject Composition.
}
\label{fig:composition_data_pipeline}
\end{figure}

\noindent \textbf{Multi-Subject Composition.} Finally, we addressed multi-subject image composition using the open-source MUSAR-Gen \citep{guo2025musar} dataset and a new synthetic data pipeline. 
As illustrated in Figure~\ref{fig:composition_data_pipeline}, we design a synthetic pipeline to construct high-quality multi-subject composition data. 
Starting from an in-house database, we combine both real and synthetic images, and generate human–object–scene lists that are further refined by large language models (LLMs) to produce natural composition instructions. 
Grounding-DINO \cite{grounding_dino} and SAM \cite{kirillov2023segment} are employed to extract object-level masks and build a mask gallery, which provides structural guidance for subsequent composition. 
Reference images of subjects and objects are synthesized by UNO-FLUX and GPT-Image1, while scene backgrounds are generated by mask-inpainting model.
The resulting target images, together with corresponding composition instructions and scene prompts, form diverse training triplets that enhance the coverage of multi-subject scenarios.
This yielded 40k multi-subject reference examples, each featuring compositions of multiple humans, objects, and complex scenes, thereby enriching the dataset’s coverage of realistic multi-entity interactions.

\section{Experiments}

\subsection{Quantitative results}

\begin{table}[t]\centering
\caption{\textbf{Quantitative Evaluation results on GenEval~\citep{ghosh2023geneval}.} $\dagger$ refer to the methods using the LLM rewriter.}
\resizebox{\textwidth}{!}{
    \begin{tabular}{l|cccccc|c}
    \toprule
    \multirow{2}{*}{\textbf{Model}} & \textbf{Single} & \textbf{Two} & \multirow{2}{*}{\textbf{Counting}} & \multirow{2}{*}{\textbf{Colors}} & \multirow{2}{*}{\textbf{Position}} & \textbf{Attribute} & \multirow{2}{*}{\textbf{Overall$\uparrow$}} \\
    & \bf Object & \bf Object & & & & \bf Binding & \\
    \midrule
    Show-o~\citep{xie2024show} & 0.95 & 0.52 & 0.49 & 0.82 & 0.11 & 0.28 &0.53 \\
    Emu3-Gen~\citep{wang2024emu3} & 0.98 & 0.71 & 0.34 & 0.81 & 0.17 &0.21 & 0.54 \\
    PixArt-$\alpha$~\citep{chen2024pixartalpha}         & 0.98          & 0.50        & 0.44     & 0.80    & 0.08     & 0.07              & 0.48     \\
    SD3 Medium~\citep{esser2024scaling}      & 0.98          & 0.74       & 0.63     & 0.67   & 0.34     & 0.36              & 0.62     \\
    FLUX.1 [Dev]~\citep{flux2024}   & 0.98 & 0.81 & 0.74  & 0.79 & 0.22   & 0.45 & 0.66     \\
    SD3.5 Large~\citep{esser2024scaling}     & 0.98          & 0.89       & 0.73     & 0.83   & 0.34     & 0.47              & 0.71     \\
    JanusFlow~\citep{ma2025janusflow} &0.97 & 0.59 & 0.45 & 0.83 & 0.53 & 0.42 & 0.63 \\
    Lumina-Image 2.0~\citep{qin2025lumina} & - & 0.87 & 0.67 & -      & - & 0.62 & 0.73     \\
    Janus-Pro-7B$^{\dagger}$~\citep{chen2025janus}     & 0.99          & 0.89       & 0.59     & 0.90    & 0.79     & 0.66              & 0.80      \\
    HiDream-I1-Full$^{\dagger}$~\citep{cai2025hidream}          & \bf 1.00             & 0.98       & 0.79     & 0.91   & 0.60      & 0.72              & 0.83     \\
    GPT-Image$^{\dagger}$~\citep{gptimage}           & 0.99          & 0.92       & 0.85     & 0.92   & 0.75     & 0.61              & 0.84     \\
    Seedream 3.0$^{\dagger}$~\citep{gao2025seedream} & 0.99 & \bf 0.96 & \bf 0.91 & 0.93 & 0.47 & 0.80 &0.84 \\
    Qwen-Image$^{\dagger}$~\citep{wu2025qwenimagetechnicalreport}  & 0.99 & 0.92 & 0.89 & 0.88 &  0.76 & 0.77 & {0.87} \\
    BAGEL$^{\dagger}$~\citep{deng2025bagel} & 0.98 & 0.95  & 0.84 & \bf0.95 & 0.78 &0.77 & \textbf{0.88} \\
    
    \midrule
    \rowcolor{lightblue}\bf {\ours}$^{\dagger}$  & 0.98 & 0.94 & 0.81 & 0.91 &  \bf 0.85 & \bf 0.79 & \bf {0.88} \\
    \bottomrule
    \end{tabular}
}
\label{tab:geneval}
\end{table}

\begin{table*}[!h]
    \centering
    \caption{\textbf{Quantitative Evaluation results on GEdit-Bench.} G\_SC is Semantic Consistency, G\_PQ is Perceptual Quality, and G\_O is Overall Score which is computed as the geometric mean of G\_SC and G\_PQ, averaged over all samples. All metrics are evaluated by GPT-4. We highlight the \textbf{best} and \underline{second-best} values for each metric.}
    \begin{tabular}{l|ccc|ccc}
    \toprule
    \multirow{2}{*}{\bf Model} & \multicolumn{3}{c|}{\bf GEdit-Bench-EN (Full set)$\uparrow$} & \multicolumn{3}{c}{\bf GEdit-Bench-CN (Full set)$\uparrow$} \\
    \cmidrule{2-4} \cmidrule{5-7}
    & G\_SC & G\_PQ & G\_O & G\_SC & G\_PQ & G\_O \\
    \midrule
    Instruct-Pix2Pix \citep{brooks2023instructpix2pix} & 3.58 & 5.49 & 3.68 & - & - & - \\
    AnyEdit~\citep{yu2025anyedit} & 3.18 & 5.82 & 3.21 & - & - & - \\
    MagicBrush~\citep{zhang2023magicbrush} & 4.68 & 5.66 & 4.52 & - & - & - \\
    UniWorld-v1~\citep{lin2025uniworldv1} & 4.93 & 7.43 & 4.85 & - & - & - \\
    OmniGen~\citep{xiao2025omnigen} & 5.96 & 5.89 & 5.06 & - & - & - \\
    OmniGen2~\citep{wu2025omnigen2} & 7.16 & 6.77 & 6.41 & - & - & - \\
    Gemini 2.0~\citep{googleGemini2} & 6.73 & 6.61 & 6.32 & 5.43 & 6.78 & 5.36 \\
    BAGEL~\citep{deng2025bagel} & 7.36 & 6.83 & 6.52 & 7.34 & 6.85 & 6.50 \\
    FLUX.1 Kontext [Pro]~\citep{labs2025flux_kontext} & 7.02 & 7.60 & 6.56 & 1.11 & 7.36 & 1.23 \\
    Step1X-Edit~\citep{liu2025step1x} & 7.66 & 7.35 & 6.97 & 7.20 & 6.87 & 6.86 \\
    GPT Image 1 [High]~\citep{gptimage} & {7.85} & \underline{7.62} & {7.53} & {7.67} & \underline{7.56} & {7.30} \\
    Qwen-Image \cite{wu2025qwenimagetechnicalreport} & \underline{8.00} & \textbf{7.86} & \underline{7.56} & \underline{7.82} & \textbf{7.79} &  \underline{7.52} \\
    \midrule
    \rowcolor{lightblue}\bf {\ours} & \textbf{8.36} & 7.37 & \textbf{7.66} & \textbf{8.39} & 7.35 & \textbf{7.65} \\
    \bottomrule
    \end{tabular}
\label{tab:gedit}
\end{table*}
We evaluate {\ours} on a comprehensive suite of benchmarks, spanning text-to-image generation, instruction-guided editing, subject-driven customization, and multi-subject composition. Specifically, we report results on GenEval, GEdit-Bench, DreamBooth, and DreamBench.

On {GenEval}, {\ours} attains an overall score of {0.88}, matching the SOTA result of unified UMM (BAGEL \citep{deng2025bagel}), as illustrated in Table \ref{tab:geneval}. Our results are reported based on Chinese prompts rewritten by DeepSeek \footnote{\url{https://chat.deepseek.com/}}. 
On GEdit-Bench, {\ours} achieves the highest overall performance in instruction-guided editing, with scores of 7.66 on the English split and 7.65 on the Chinese split. These results surpass Qwen-Image (7.56 / 7.52) and GPT-Image (7.53 / 7.30), as shown in Table \ref{tab:gedit}.
We note that the Perceptual Quality score exhibits some shortcomings, primarily due to the lack of a reinforcement learning or supervised fine-tuning stage designed to enhance generation quality or photorealism. We leave this exploration in future work.
For subject-driven generation on DreamBooth, {\ours} establishes new state-of-the-art results with {DINO 0.786} and {CLIP-I 0.858}, significantly outperforming Metaquery (0.737 / 0.851) and UNO-FLUX (0.760 / 0.835), though with a slightly lower CLIP-T (0.307 vs. OmniGen’s 0.315), as shown in Table \ref{tab:single_subject}. 
In Table~\ref{tab:multi_subject}, {\ours} achieves the best CLIP-T score ({0.336}) alongside competitive DINO (0.532) and CLIP-I results (0.731), on the {multi-subject composition} benchmark DreamBench.

\subsection{Qualitative Results}  
We also provide qualitative comparisons across all task categories, including text-to-image generation, instruction editing, and customized generation, under both Chinese and English prompts. Representative examples are shown in Figure~\ref{fig:cover_img}.

\begin{table}[t]
    \centering
    \caption{\textbf{Quantitative results for single-subject driven generation on Dreambooth.} We highlight the \textbf{best} and \underline{second-best} values.}
    \label{tab:single_subject}
    \begin{tabular}{lccc}
    \toprule
    \textbf{Method} & \textbf{DINO} $\uparrow$ & \textbf{CLIP-I} $\uparrow$ & \textbf{CLIP-T} $\uparrow$ \\ 
    \midrule
    \rowcolor{gray!20}
    \multicolumn{1}{l}{\textsl{Tuning-free}}  & \multicolumn{3}{l}{} \\
    Textual Inversion \cite{gal2022image} & 0.569 & 0.780 & 0.255\\
    DreamBooth \cite{ruiz2023dreambooth} &0.668  & 0.803 & 0.305 \\
    BLIP-Diffusion \cite{li2023blip} & 0.670 & 0.805 & 0.302\\
    \midrule
    \rowcolor{gray!20}
    \multicolumn{1}{l}{\textsl{Specialist Models}}  & \multicolumn{3}{l}{} \\
    ELITE \cite{wei2023elite}& 0.647  & 0.772 & 0.296\\
    Re-Imagen \cite{chen2022re}& 0.600  & 0.740 & 0.270\\
    OminiControl \cite{tan2024ominicontrol}& 0.684  & 0.799 & 0.312\\
    FLUX.1 IP-Adapter \cite{flux2024} & 0.582  & 0.820 & 0.288\\
    UNO-FLUX \cite{wu2025uno_flux} & \underline{{0.760}}  & {{0.835}} & {0.304} \\
    \midrule
    \rowcolor{gray!20}
    \multicolumn{1}{l}{\textsl{Generalist Models}}  & \multicolumn{3}{l}{} \\
    OmniGen \cite{xiao2024omnigen}& 0.693  & 0.801 & \textbf{0.315}\\
    Metaquery \citep{metaquery} & 0.737  & \underline{0.851} & {0.301}\\
    \rowcolor{lightblue}\bf {\ours}  & \textbf{0.786} & \textbf{0.858} & \underline{0.307} \\ 
    \bottomrule
    \end{tabular}
\end{table}
    
\begin{table}[t]
    \centering
    \caption{\textbf{Quantitative results for multi-subject driven generation on Dreambench.} We highlight the \textbf{best} and \underline{second-best} values for each metric.}
    \label{tab:multi_subject}
    \begin{tabular}{lcccc}
    \toprule
    \textbf{Method} & \textbf{DINO} $\uparrow$ & \textbf{CLIP-I} $\uparrow$ & \textbf{CLIP-T} $\uparrow$ \\ 
    \midrule
    \rowcolor{gray!20}
    \multicolumn{1}{l}{\textsl{Tuning-free}}  & \multicolumn{3}{l}{} \\
    DreamBooth \cite{ruiz2023dreambooth} &0.430  & 0.695 & 0.308 \\
    BLIP-Diffusion \cite{li2023blip} & 0.464 & 0.698 & 0.300\\
    \midrule
    \rowcolor{gray!20}
    \multicolumn{1}{l}{\textsl{Specialist Models}}  & \multicolumn{3}{l}{} \\
    Subject Diffusion \cite{ma2024subject}& 0.506  & 0.696 & 0.310\\
    MIP-Adapter \cite{huang2024resolving}& 0.482  & {0.726} & 0.311\\
    MS-Diffusion \cite{wang2025msdiffusion}& {0.525}  & {0.726} & 0.319\\
    {UNO-FLUX} \cite{wu2025uno_flux} & {\textbf{0.542}}  & {\textbf{0.733}} & {0.322}\\
    \midrule
    \rowcolor{gray!20}
    \multicolumn{1}{l}{\textsl{Generalist Models}}  & \multicolumn{3}{l}{} \\
    OmniGen \cite{xiao2024omnigen}& 0.511  & 0.722 & {\underline{0.331}}\\
    \rowcolor{lightblue}\bf {\ours} & \underline{{0.532}}  & \underline{{0.731}} & \textbf{0.336}\\
    \bottomrule
    \end{tabular}
\end{table}

\subsection{Shifted RoPE}  



We further examine the effect of the proposed shifted 2D-RoPE mechanism for handling reference images. With \textit{source} input images, the model tends to preserve the pixel-level fidelity of the input, producing faithful reconstructions. In contrast, with \textit{reference} input images, the model emphasizes instruction following and generalization, maintaining subject identity while generating more diverse outputs. Comparative results on the DreamBooth benchmark using source versus reference images are reported in {Table~\ref{tab:rope}.}

\subsection{Query–Kontext Convergence}

\begin{figure}[htbp] 
\centering 

\begin{minipage}[t]{0.46\textwidth}
    \centering 

    \captionof{table}{The comparison between the shifted RoPE on $source$ or $reference$ image.}
    
    \label{tab:rope}
    \begin{adjustbox}{width=\linewidth}
    \begin{tabular}{lccc}
        \toprule
        \textbf{Method} & \textbf{DINO} $\uparrow$ & \textbf{CLIP-I} $\uparrow$ & \textbf{CLIP-T} $\uparrow$ \\
        \midrule
        w/ ${img}_{src}$ & 0.865 & 0.914 & 0.289 \\
        w/ ${img}_{ref}$ & 0.786 & 0.858 & 0.307 \\
        \bottomrule
    \end{tabular}
    \end{adjustbox}

    \vspace{10pt} 

    \captionof{table}{Ablations on positional encoding, number of reference images, and LoRA ranks.}
    \label{tab:ablations}
    \begin{adjustbox}{width=\linewidth}
    \begin{tabular}{lcccc}
        \toprule
        \bf Setting & \bf DINO & \bf CLIP-I & \bf CLIP-T \\
        \midrule
        LoRA $r{=}64$ & 0.752 & 0.841 & 0.298 \\
        LoRA $r{=}128$ & 0.786 & 0.858 & 0.307 \\
        LoRA $r{=}256$ & 0.777 & 0.834 & 0.311 \\
        \bottomrule
    \end{tabular}
    \end{adjustbox}

\end{minipage} 
\hfill 
\begin{minipage}[t]{0.5\textwidth}
    \centering 
    \adjustbox{valign=t}{
        \includegraphics[width=1.0\linewidth]{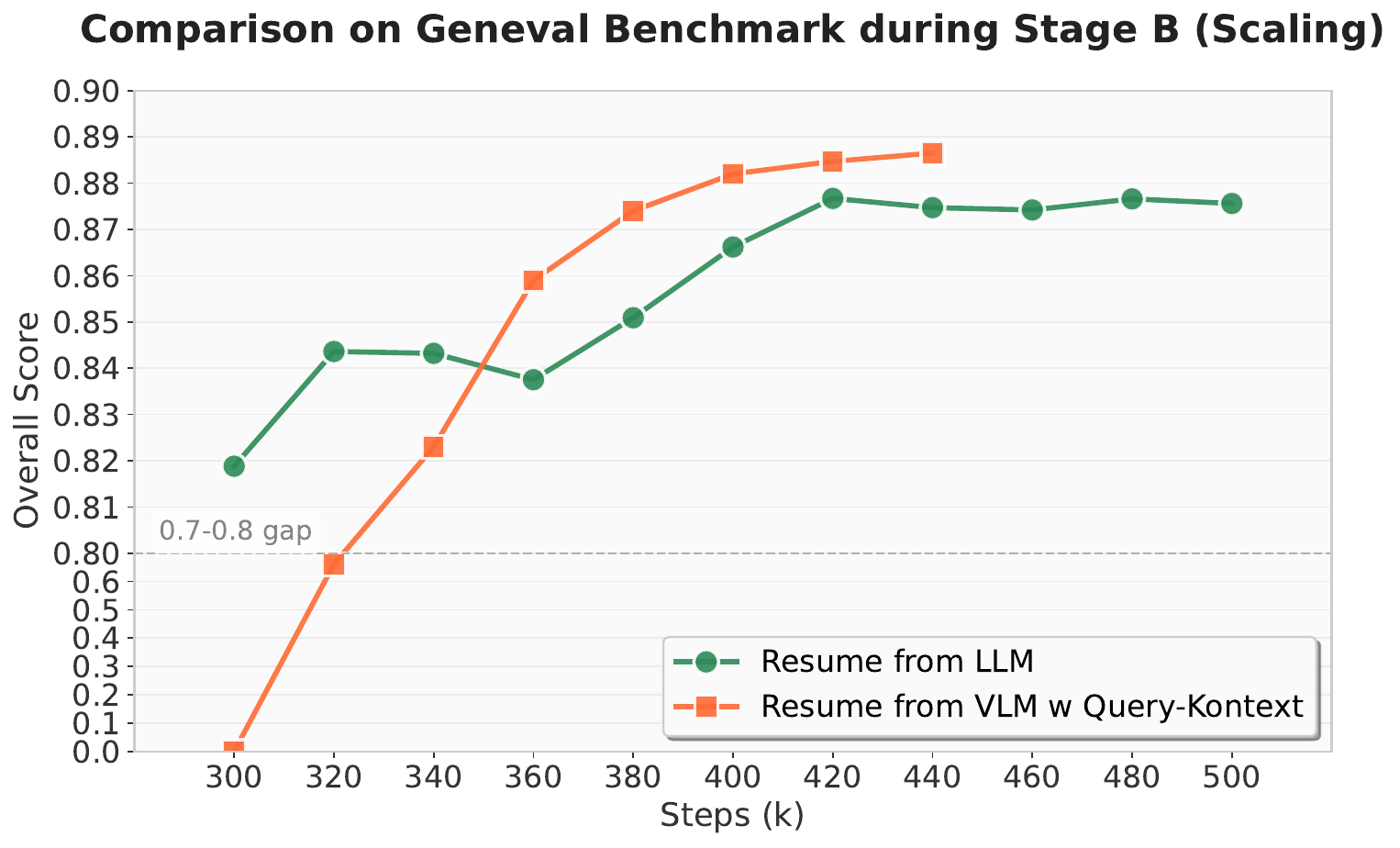}
    }
    
    \captionof{figure}{\textbf{Convergence validation of Query–Kontext.} Comparison on our in-house MMDiT between VLM re-alignment with Query–Kontext and LLM-based resumption.}
    \label{fig:geneval_abl}

\end{minipage} 

\end{figure}

In Stage~2, we analyze the convergence behavior of the diffusion model when conditioned on two settings: (i) text-only embeddings from an LLM and (ii) the mixed conditioning from both text tokens and Query–Kontext tokens generated by our fine-tuned VLM. We observe that replacing the LLM with our VLM leads to faster alignment of the diffusion model and produces superior visual results compared to the LLM-conditioned baseline, as shown in Figure ~\ref{fig:geneval_abl}. This demonstrates that decoupling multimodal reasoning from visual generation via Query–Kontext not only accelerates convergence but also unleashes the full potential of both the VLM and the diffusion model.

\subsection{LoRA Rank}
\label{app:ablation_lora}
We evaluate LoRA ranks $\{64,128,256\}$ on both the diffusion model and the MLLM adapters, observing faster convergence at higher ranks with marginal quality gains beyond $r{=}128$.


\section{Related Work}



\subsection{Instruction-based Image Editing}
Early diffusion-based editing splits into \textit{training-free} and \textit{training-based} approaches. Training-free methods manipulate the denoising trajectory via inversion~\cite{meng2021sdedit, mokady2023null, kawar2023imagic, couairon2022diffedit, wallace2023edict, chen2023photoverse} or attention control~\cite{cao2023masactrl, hertz2022prompt, parmar2023zero} and require no additional training, but often struggle with fine-grained instruction fidelity and identity preservation. InstructPix2Pix~\cite{brooks2023instructpix2pix} pioneered training-based approaches~\cite{chen2023photoverse, zhao2024ultraedit, li2024brushedit, yu2025anyedit, zhang2025context} by finetuning a pretrained diffusion backbone on curated (image, instruction, edited-image) triplets, yielding stronger instruction following and higher fidelity. 
More recently, a trend towards tighter integration of \textit{multimodal} understanding and generation has emerged to empower more complex editing instructions. 
Works like SmartEdit~\cite{huang2024smartedit} and Step1X‑Edit~\cite{liu2025step1x} leverage MLLM latent representations to guide structured or latent-conditioned editing, ACE~\cite{han2024ace}, ACE++~\cite{mao2025ace++} and FLUX.1 Kontext~\cite{labs2025flux_kontext} integrate text and image context for instruction-guided editing. UniVG~\cite{fu2025univg}, SeedEdit 3.0~\cite{wang2025seededit}, and Qwen‑Image~\cite{wu2025qwenimagetechnicalreport} demonstrate generalist architectures capable of diverse tasks while preserving identity and fidelity.

\subsection{Unified Multimodal Models}

Unified Multimodal Models (UMMs) have recently attracted significant attention for their ability to unify both understanding and generation within a single architecture. Existing approaches can be broadly categorized into two strategies.
The first strategy develops \textit{native} UMMs~\cite{chameleon, transfusion, xie2024show, xie2025show, janus2024, januspro2025, janusflow2024, emu3, tong2024metamorph, deng2025bagel}, which are trained to fuse multimodal understanding and generation capabilities at the early stage, usually involving autoregressive or diffusion modeling. While conceptually elegant, they often present considerable challenges in training and scaling.
The second strategy \textit{assembles} unified frameworks~\cite{metaquery, wu2025qwenimagetechnicalreport, chen2025blip3, labs2025flux_kontext, liu2025step1xeditpracticalframeworkgeneral, chen2025unireal, fu2025univg, chen2025multimodal} by coupling existing vision-language models (VLMs)\cite{qwen2.5-vl, qwen2vl} for understanding with powerful diffusion-based generators\cite{SD3, dit, flux}. This is typically achieved through 
learnable tokens or tuning adapters. Our work builds on this line of research, introducing a more refined mechanism for cross-modal representation fusion and controllable generation.

\subsection{Editing Data Curation}
High-quality and diverse datasets of ⟨original image, instruction, edited image⟩ triplets are fundamental for training powerful editing models. MagicBrush~\cite{zhang2023magicbrush} represents the manual annotation approach. InstructPix2Pix~\cite{brooks2023instructpix2pix} pioneered data synthesis by using GPT-3~\cite{gpt3} and Prompt-to-Prompt~\cite{hertz2022prompt}. To improve quality, HIVE~\cite{zhang2024hive} introduced human feedback for quality assessment and training. HQ-Edit~\cite{hui2024hqedit} and UltraEdit~\cite{zhao2024ultraedit} scaled up dataset size and difficulty using more powerful models like GPT-4V~\cite{gpt4v} and DALL-E 3~\cite{dalle3}, along with fine-grained annotations. SEED-Data-Edit~\cite{ge2024seed} enhances diversity through re-generation and re-annotation techniques, while SeedEdit 3.0~\cite{wang2025seededit} systematically upgrades both data sources and data merging. More recently, NHR-Edit~\cite{kuprashevich2025nohumansrequired} automates the mining of high-quality triplets from powerful open-sourced generative models like FLUX~\cite{flux}, reducing manual effort and improving data realism.

\section{Discussion}
\label{sec:disscusion}
\noindent \textbf{Economical Alignment between VLM and Diffusion Model.}  
\ours\ builds on a powerful VLM and an MMDiT-based diffusion model, leveraging the strengths of each to construct a unified multimodal-to-image generation system. The training process was conducted on 192 NVIDIA H100 GPUs (80GB), which amounts to roughly 10\% of the computational resources typically required to train a large-scale diffusion model from scratch (e.g., Qwen-Image) or an integrated multimodal transformer (e.g., BAGEL). This economical alignment allows us to allocate resources more effectively, focusing on higher-level and underexplored post-training tasks such as multi-subject composition, multi-image generation, and interleaved text–image generation.

\noindent \textbf{Scaling of the Diffusion Model.}  
By decoupling multimodal generative reasoning in the VLM from high-fidelity visual synthesis in the diffusion model, our framework enables independent exploration on the scaling laws of each component. 
This separation is crucial, as VLMs and diffusion models often exhibit competing capacity requirements and benefit from different parameter budgets. 
In Stage~2, we attempted alignment with in-house diffusion backbones of varying sizes (0.9B, 4B, and 10B parameters). 
However, alignment was not always successful, particularly when employing a lightweight connector to bridge a heavy and frozen diffusion model (e.g., 10B parameters). 
To mitigate this issue, we unfroze the diffusion model parameters during Stage~2 training, thereby avoiding an intensive grid search over connector hyperparameters. 
Investigating the scaling laws governing the connector remains an important direction for future work.

\section{Conclusion}

In this work, we introduced \ours, an economical unified multimodal-to-image framework that decouples multimodal generative reasoning (handled by the VLM) from high-fidelity rendering (handled by the diffusion model). To fully harness the potential of both components, we proposed a three-stage progressive training strategy that progressively aligns the VLM with increasingly capable diffusion generators while amplifying their complementary strengths. In addition, we curated a multimodal reference-to-image dataset mixture spanning real, synthetic, and carefully filtered open-source data. Extensive experiments demonstrate that our framework achieves competitive performance across diverse tasks, including image generation, instruction editing, customized subject synthesis, and multi-subject composition.

\clearpage
\bibliographystyle{plainnat}  
\bibliography{Query-Kontext}          



\end{document}